\documentclass[runningheads]{waica}
\usepackage[T1]{fontenc}
\usepackage{graphicx}
\usepackage{amsmath}
\usepackage{booktabs}
\graphicspath{{figures/}}

\newcommand{\system}{\textsc{InflationAgent}}

\begin{document}

\title{Not All Tokens Are Equal: Inflation-Aware Routing for Agentic LLM Systems}
\titlerunning{Inflation-Aware Routing for Agentic LLM Systems}

% \author{Heming Fu\inst{1} \and Shan Lin\inst{1} \and Guojun Xiong\inst{2}\thanks{Corresponding author.}}
% \authorrunning{H. Fu et al.}

% \institute{Stony Brook University, Stony Brook NY 11794, USA\\
% \email{\{heming.fu, shan.x.lin\}@stonybrook.edu} \and
% Shanghai Jiao Tong University, Shanghai, China\\
% \email{gjxiong@sjtu.edu.cn}}

\author{
Heming Fu\inst{1} \and
Shan Lin\inst{1} \and
Qianqian Xie\inst{2} \and
Guojun Xiong\inst{3}
}
\authorrunning{H. Fu et al.}

\institute{
Stony Brook University
\email{\{heming.fu, shan.x.lin\}@stonybrook.edu}
\and
Wuhan University
\email{\{xieq@whu.edu.cn\}}
\and
Shanghai Jiao Tong University
\email{\{gjxiong@sjtu.edu.cn\}}
}

\maketitle

\begin{abstract}
When a language model fails to answer a query on the first attempt, an agentic system retries,
consuming additional tokens each time. This retry overhead creates a gap between what a model's
per-token price implies and what a full workflow actually costs. We call this gap \emph{token
inflation} and define it as the ratio of true workflow cost to single-call cost. Systems like
FrugalGPT~\cite{frugalgpt} route based on the latter, which can underestimate real cost by
more than $2\times$ on difficult tasks.
We address this with \system{}, a four-stage router that (1) measures token inflation
systematically across model tiers and task types, finding inflation as high as $4.25\times$
for a 7B model on multi-hop question answering; (2) introduces CoT Branching Entropy (CBE),
a pre-execution difficulty signal computed entirely from local inference, which predicts high
inflation with AUROC 0.887; and (3) selects models by maximizing a Semantic Exchange Rate
(SER) that divides expected accuracy by predicted true cost, with a fresh-escalation policy
that discards failed chains before routing to a stronger model. On GSM8K under a fixed budget,
\system{} achieves 94.7\% accuracy versus 91.0\% for FrugalGPT while using 31\% fewer tokens,
and we show that forwarding a failed reasoning chain to GPT-4o reduces its accuracy by up to
34.8 percentage points, validating the fresh-escalation design.

\end{abstract}

\section{Introduction}

Modern agentic systems built on large language models rarely issue a single call and move
on. When a model returns a wrong answer or an incomplete reasoning chain, the agent
retries~\cite{react,reflexion}. ReAct-style agents re-issue queries after tool errors;
self-consistency methods~\cite{selfconsistency} sample multiple chains; code-generation
agents loop until tests pass~\cite{agentbench}. The cost of these retries is real but
invisible to the routing decision.

Consider a concrete example. A user asks a pipeline to answer: ``Who directed the film
whose soundtrack was composed by the artist who collaborated with Michael Jackson on
Thriller?'' A 7B local model might attempt this five times and still fail, consuming roughly
five times the tokens a cost estimator would predict. A GPT-4o call answers correctly on the
first try. Measured by actual workflow cost, the ``cheap'' option is not cheap at all.

Existing routing systems~\cite{frugalgpt,llmrouter,routerbench} estimate the cost of a query
as the price per token multiplied by the expected output length of a single call. When the
model fails on the first attempt, this estimate can be off by a factor of two to five. We
call this discrepancy \emph{token inflation}: the ratio between what a workflow actually costs
and what a single-call estimate predicts.

This paper introduces \system{}, a routing system that makes three contributions.
\textbf{Measurement}: we instrument an agentic retry harness across two reasoning tasks and
three model tiers to show that inflation varies dramatically by task type, reaching $4.25\times$
for a 7B model on multi-hop QA versus $1.31\times$ for GPT-4o.
\textbf{Prediction}: we introduce CoT Branching Entropy (CBE), a pre-execution difficulty
signal that estimates retry likelihood from a small sample of local reasoning chains, with
no API cost.
\textbf{Routing}: we define the Semantic Exchange Rate (SER) as expected accuracy divided by
predicted true cost, route to the model with the highest SER, and escalate with a fresh
prompt when the chosen model inflates beyond its prediction.

\begin{figure}[t]
  \centering
  \includegraphics[width=\textwidth]{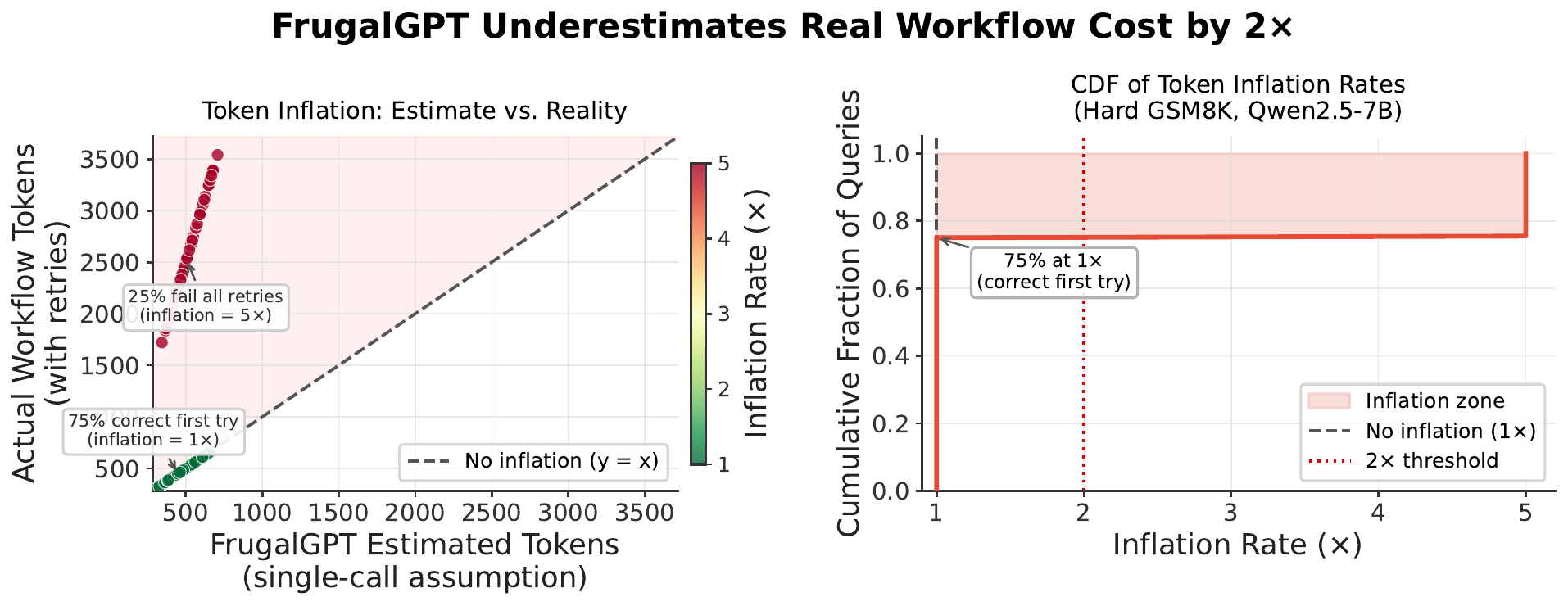}
  \caption{Token inflation on 200 hard GSM8K queries (Qwen2.5-7B, max 5 retries).
    \textbf{Left:} Each point is one query. The x-axis is the FrugalGPT single-call cost
    estimate; the y-axis is the actual workflow cost. Color encodes inflation rate from
    green ($1\times$, answered correctly on first try) to red ($5\times$, all retries
    exhausted). Points above the diagonal have inflated cost.
    \textbf{Right:} CDF of inflation rates. The distribution is bimodal: 75\% of hard
    queries succeed on the first attempt ($1\times$ cost) and 25\% fail all five retries
    ($5\times$ cost), making the mean $2\times$ and causing FrugalGPT to underestimate
    actual workflow cost by 110\%.}
  \label{fig:crack}
\end{figure}

Figure~\ref{fig:crack} illustrates the core motivation: each point is one query, plotted by
its FrugalGPT estimated token count versus its actual workflow token count. Points above the
diagonal represent the hidden cost that routing systems currently ignore.

\section{Related Work}

\paragraph{LLM cost routing and cascades.}
A growing body of work reduces inference cost by routing easy queries to smaller models.
FrugalGPT~\cite{frugalgpt} learns a cascade policy that stops escalating once a response
passes a quality threshold. RouteLLM~\cite{routellm} trains a router on human preference
data; Hybrid LLM~\cite{llmrouter} routes by query complexity; RouterBench~\cite{routerbench}
benchmarks routing policies across model families; LLM-Blender~\cite{llmblender} fuses outputs
from multiple models; language model cascades~\cite{cascades} query increasingly capable models
until one produces a confident answer. All of these systems estimate routing cost as the price
per token of a single call, which is accurate only when the pipeline executes the chosen model
once. In agentic settings with retries, the single-call estimate systematically understates
the true cost. Our work models this retry overhead explicitly and incorporates it into the
routing objective.

\paragraph{Uncertainty estimation for language models.}
Self-consistency~\cite{selfconsistency} generates multiple completions and uses answer
agreement as an implicit confidence measure. Semantic uncertainty~\cite{uncertainty} clusters
semantically equivalent answers before measuring entropy, correcting for paraphrase variation.
Kadavath et al.~\cite{llmscore} show that large models can produce calibrated self-assessments
when prompted appropriately. CoT Branching Entropy (CBE) is related to self-consistency but
serves a different purpose: rather than aggregating answers for final output, we use the
entropy of a small chain sample as a \emph{pre-execution} proxy for retry probability.
The key design choice is that CBE is computed locally, adding no API cost to the routing
decision.

\paragraph{Agentic systems and multi-step inference.}
Chain-of-thought prompting~\cite{cot} improves reasoning by eliciting intermediate steps.
ReAct~\cite{react} interleaves reasoning and tool use into inspectable agent trajectories.
Reflexion~\cite{reflexion} adds verbal self-critique to allow agents to revise failed plans.
AgentBench~\cite{agentbench} evaluates LLM agents across diverse task environments.
This work establishes that retry loops are standard practice, but does not address their
cost implications. When retries are common, the expected token cost can far exceed a single
call, and our work closes this gap by routing in a way that accounts for it.

\section{Methodology}

\subsection{Problem Formulation}

We consider an agentic pipeline that selects one model $m$ from a candidate set $\mathcal{M}$
and executes it with up to $R$ retry attempts on query $x$. Let $T_k(m,x)$ denote the token
count of the $k$-th call. The true workflow cost is
\begin{equation}
  C_{\text{true}}(m, x) = \sum_{k=1}^{k^*} T_k(m, x),
\end{equation}
where $k^*$ is the index of the first successful attempt ($k^*=R$ if the model never succeeds).
Existing routers approximate this as $C_{\text{direct}}(m,x) = T_1(m,x)$, the single-call cost.
\textbf{Token inflation} is the ratio
\begin{equation}
  \text{Inflation}(m, x) = C_{\text{true}}(m, x) \;/\; C_{\text{direct}}(m, x).
\end{equation}
A model that always succeeds first has inflation $1\times$; one that exhausts all $R$ retries
reaches $R\times$. Routing systems that optimize for $C_{\text{direct}}$ implicitly assume
this ratio is always 1, an assumption that fails badly on difficult multi-hop queries.

We define the utility of model $m$ as its expected accuracy $U(m)\in[0,1]$. The routing
objective is to maximize accuracy per unit of true cost. We call this the
\textbf{Semantic Exchange Rate}:
\begin{equation}
  \mathrm{SER}(m, x) = U(m) \;/\; C_{\text{true}}(m, x), \qquad
  m^* = \arg\max_{m \in \mathcal{M}} \mathrm{SER}(m, x).
\end{equation}
SER rewards models that are accurate \emph{and} reliable: a high-accuracy model with severe
inflation can score lower than a slightly less accurate one that rarely retries. Because
$C_{\text{true}}$ is not observable before execution, the system must predict it, which is
the role of the two components described below.

\subsection{CoT Branching Entropy}
\label{sec:cbe}

We use the \emph{consistency} of a model's reasoning across independent samples as a proxy
for query difficulty. An easy query produces the same answer from nearly every chain; a hard
query produces divergent paths and conflicting answers. Formally, we sample $K$ independent
reasoning chains from the local model, extract the final answer from each, and let $p_c$
be the empirical frequency of answer $c$. The \textbf{CoT Branching Entropy} (CBE) is the
Shannon entropy~\cite{shannon} of this answer distribution:
\begin{equation}
  \mathrm{CBE}(x) = -\sum_{c} p_c \log_2 p_c.
\end{equation}
CBE is always computed using the local model regardless of which tier will be deployed,
so it requires no API calls. A query that confuses a small model is genuinely ambiguous or
multi-step, and this difficulty predicts retry probability across all tiers, as we validate
in Section~\ref{sec:exp_predictor}.

\subsection{Inflation Prediction and the \system{} Pipeline}
\label{sec:iar}

\begin{figure}[t]
  \centering
  \includegraphics[width=\textwidth]{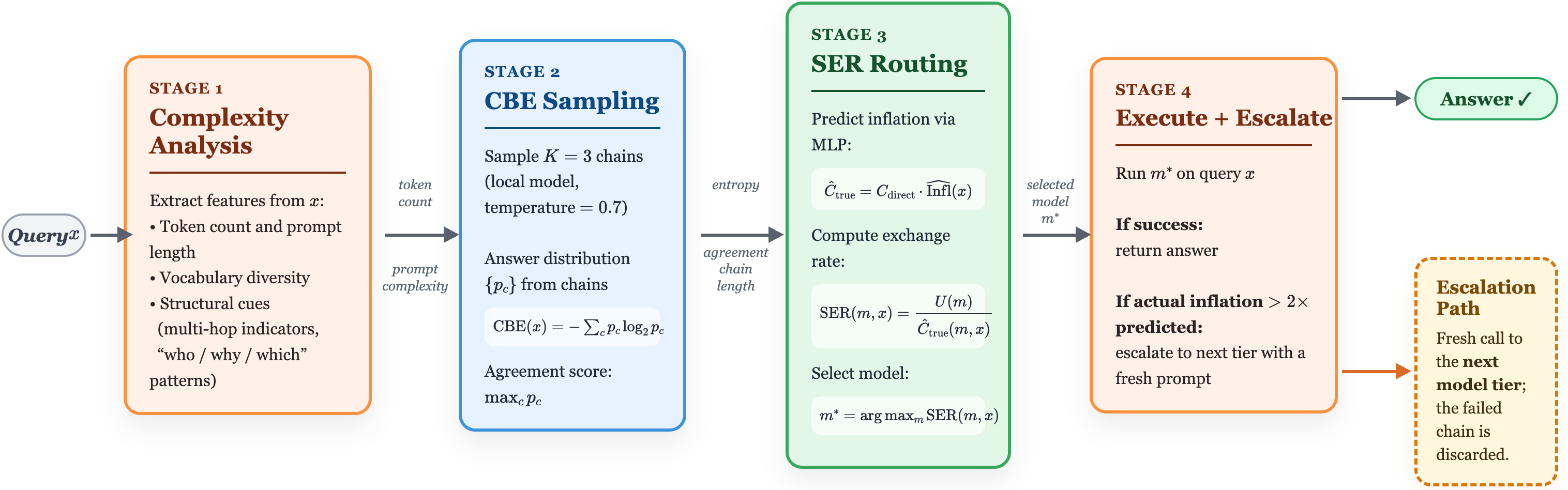}
  \caption{\system{} four-stage pipeline. \textbf{Stage 1} (orange) extracts structural
    features from the incoming query: token count, vocabulary diversity, and cues that suggest
    multi-hop reasoning (e.g., nested ``who/which'' constructions). \textbf{Stage 2} (blue)
    samples $K$ short reasoning chains from the local model and computes the CoT Branching
    Entropy $\mathrm{CBE}(x)=-\sum_c p_c\log_2 p_c$ over the resulting answer distribution,
    along with the agreement score $\max_c p_c$. Both stages run on local hardware with no
    API cost. \textbf{Stage 3} (green) feeds these features into a lightweight MLP that
    predicts $\widehat{\text{Inflation}}(x)$, forms the estimated true cost
    $\hat{C}_{\text{true}}=C_{\text{direct}}\cdot\widehat{\text{Inflation}}$, and selects
    the model with the highest Semantic Exchange Rate
    $m^*=\arg\max_m U(m)/\hat{C}_{\text{true}}(m,x)$. \textbf{Stage 4} (orange) executes
    $m^*$ on query $x$. On success it returns the answer directly (upper path). If actual
    inflation exceeds $2\times$ the prediction, \system{} escalates to the next model tier
    with a \emph{fresh} prompt (lower path, orange arrow): the failed reasoning chain is
    discarded rather than forwarded. This design is empirically validated in
    Section~\ref{sec:exp_contamination}.}
  \label{fig:iar_pipeline}
\end{figure}

CBE alone does not capture all sources of difficulty, so \system{} trains a lightweight MLP
on four features: CBE, agreement score ($\max_c p_c$), average chain length, and prompt token
count. The MLP outputs $\widehat{\text{Inflation}}(x)$, which is then used in Stage 3 to
compute the estimated true cost
$\hat{C}_{\text{true}}(m,x) = C_{\text{direct}}(m,x) \cdot \widehat{\text{Inflation}}(x)$
and select the highest-SER model. Figure~\ref{fig:iar_pipeline} shows the full four-stage
pipeline. Stages 1 and 2 run on local hardware at no API cost; Stage 3 is a single forward
pass through the MLP; Stage 4 applies a fresh-escalation policy when observed inflation
exceeds $2\times$ the prediction, discarding the failed context before calling a stronger
model. We assign compute units of $1\times$, $3\times$, and $52\times$ to the small, medium,
and large tiers respectively, calibrated to reflect token pricing and typical output lengths.

\section{Evaluation}

\subsection{Experimental Setup}
\label{sec:setup}

\paragraph{Models and datasets.}
We evaluate three model tiers: Qwen2.5-7B-Instruct~\cite{qwen25} (small, local RTX 4090),
GPT-4o-mini (medium), and GPT-4o~\cite{gpt4} (large), the latter two via the OpenRouter API.
We use GSM8K~\cite{gsm8k}, a collection of multi-step arithmetic problems with unique
numerical answers, and HotpotQA~\cite{hotpotqa}, which requires chaining two or more
retrieval and reasoning steps to produce a short factual answer. For HotpotQA we provide
only three candidate passages per query (rather than the full ten), intentionally creating
a harder setting where inflation is most pronounced. CBE sampling uses $K=3$ chains at
temperature $\tau=0.7$; the first execution attempt uses greedy decoding ($\tau=0$) and
retries use $\tau=0.7$ to avoid deterministic repetition.

\paragraph{Retry protocol and baselines.}
All agents follow a retry-until-correct protocol with $R=5$ maximum attempts. We compare
\system{} against: \textit{All-Small} (always use the local model); \textit{FrugalGPT}~\cite{frugalgpt}
(cascade by query length using $C_{\text{direct}}$ as cost estimate); and
\textit{Confidence Escalation} (escalate when CBE is high, but without an inflation model
or fresh-call semantics).

\subsection{Token Inflation Across Models and Tasks}
\label{sec:exp_inflation}

\begin{figure}[t]
  \centering
  \includegraphics[width=\textwidth]{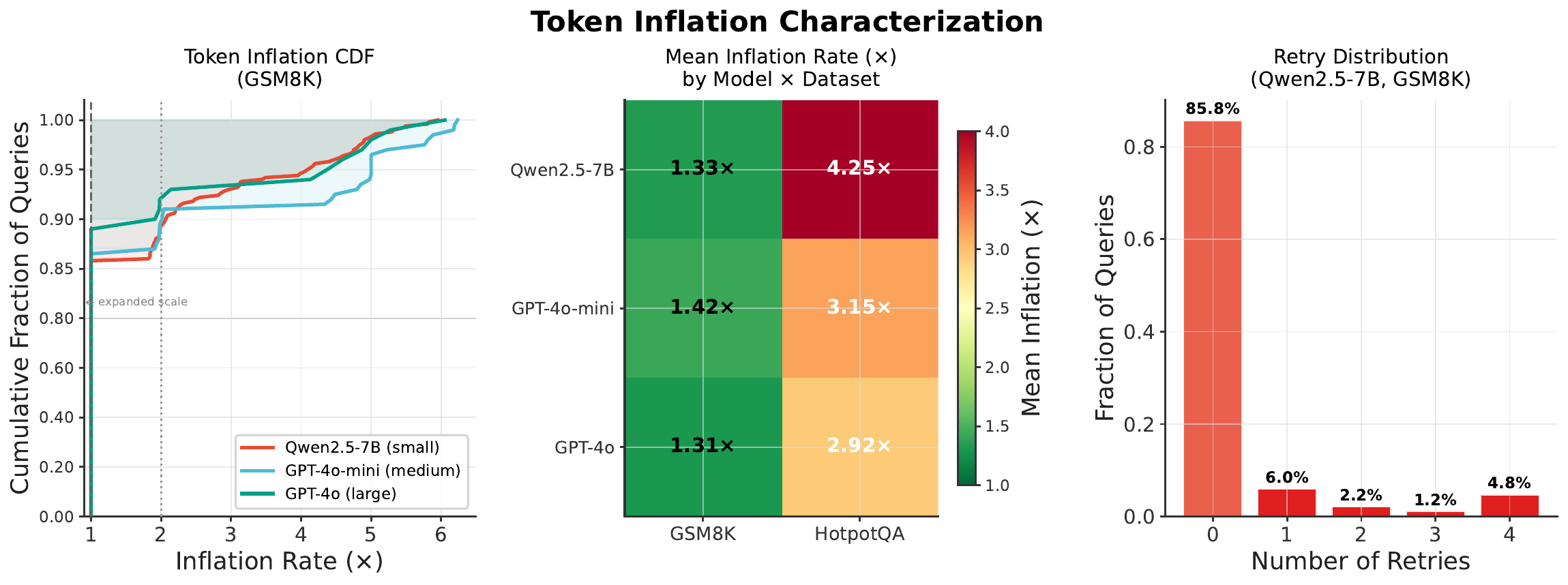}
  \caption{Inflation characterization across tiers and datasets. \textbf{Left:} CDF of
    inflation rates on GSM8K (Qwen2.5-7B in blue, GPT-4o-mini in orange, GPT-4o in green);
    the y-axis is nonlinear to expose differences among the 80--100\% of queries that
    succeed quickly. \textbf{Center:} Mean inflation heatmap (green = $1\times$, red =
    $\ge4\times$) over the full model-by-dataset matrix. \textbf{Right:} Retry count
    distribution for Qwen2.5-7B on GSM8K.}
  \label{fig:inflation}
\end{figure}

Table~\ref{tab:inflation} and Figure~\ref{fig:inflation} summarize inflation across all
tier-dataset combinations. On GSM8K, all three tiers inflate modestly (1.31--1.42$\times$),
reflecting that arithmetic reasoning is within the competence of every model tier tested.
HotpotQA tells a different story: the small model inflates by $4.25\times$, meaning the
average query consumes four times more tokens than a single-call estimate would predict.
The medium and large tiers fare better (3.15$\times$ and 2.92$\times$) but still inflate
substantially. On multi-hop queries, the small model is both expensive (80\% of queries
exceed $2\times$ inflation) and inaccurate (21.4\% task accuracy), making it a poor choice
regardless of its nominal per-token price.

\begin{table}[t]
  \centering
  \caption{Mean token inflation and task accuracy across model tiers and datasets. ``High
    Infl.'' is the fraction of queries with inflation $\ge 2\times$.}
  \label{tab:inflation}
  \small
  \begin{tabular}{lcccccc}
    \toprule
    \textbf{Model} & \multicolumn{2}{c}{\textbf{GSM8K}} & \multicolumn{2}{c}{\textbf{HotpotQA}} & \multicolumn{2}{c}{\textbf{High Infl.\ ($\ge$2$\times$)}} \\
    & Infl. & Acc. & Infl. & Acc. & GSM8K & HotpotQA \\
    \midrule
    Qwen2.5-7B  & 1.33$\times$ & 95.4\% & 4.25$\times$ & 21.4\% & 10.8\% & 80.0\% \\
    GPT-4o-mini & 1.42$\times$ & 91.0\% & 3.15$\times$ & 49.0\% & 10.5\% & 55.0\% \\
    GPT-4o      & 1.31$\times$ & 93.0\% & 2.92$\times$ & 53.0\% &  8.0\% & 48.0\% \\
    \bottomrule
  \end{tabular}
\end{table}

\subsection{Predicting Inflation Before Execution}
\label{sec:exp_predictor}

\begin{figure}[t]
  \centering
  \includegraphics[width=\textwidth]{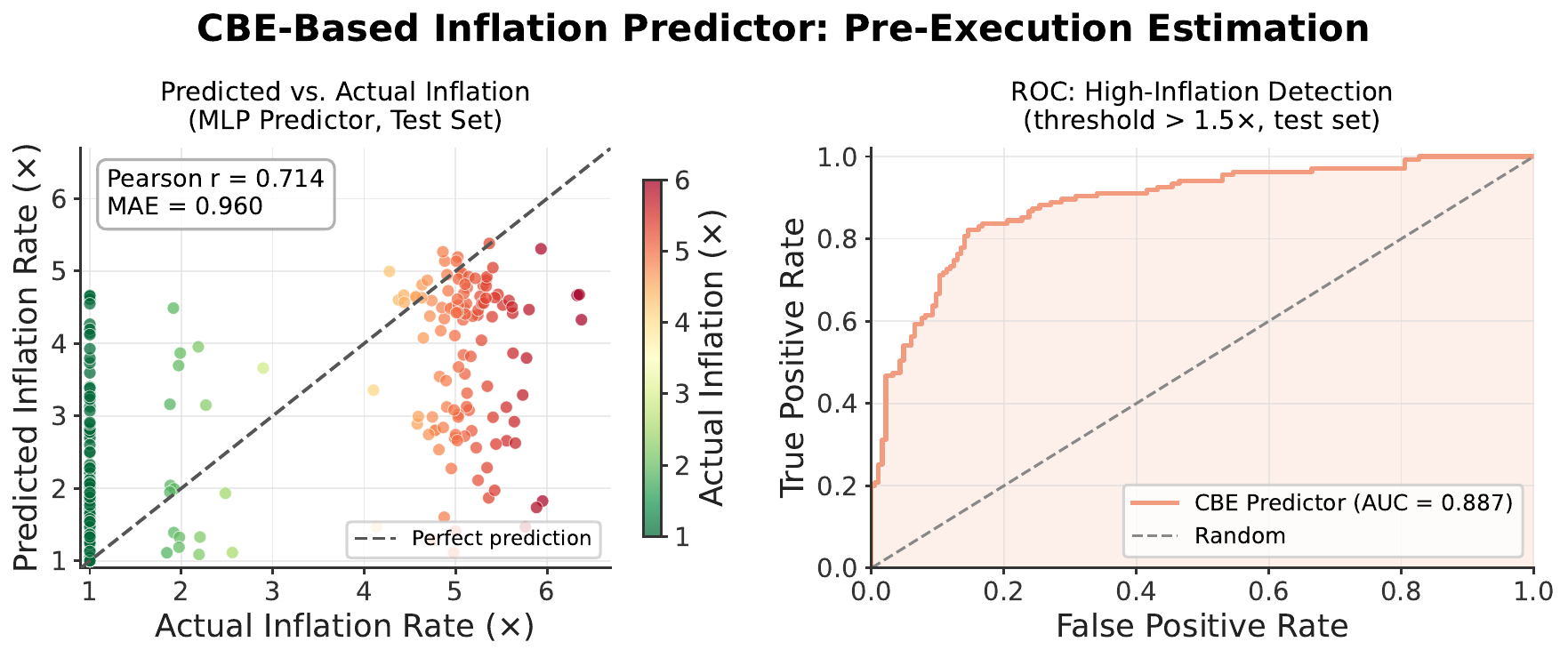}
  \caption{Inflation predictor evaluation. \textbf{Left:} Predicted vs.\ actual inflation
    on the test set; color encodes dataset (GSM8K vs.\ HotpotQA). The diagonal is perfect
    prediction. \textbf{Center:} ROC curve for identifying high-inflation queries
    ($\ge 2\times$), AUROC = 0.887. \textbf{Right:} CBE distribution split by high- vs.\
    low-inflation outcome, showing high-entropy queries are substantially more likely to
    inflate.}
  \label{fig:predictor}
\end{figure}

We train the MLP inflation predictor on 80\% of the measured data and evaluate on the
remaining 20\%. Figure~\ref{fig:predictor} and Table~\ref{tab:predictor} show the results.
The predictor achieves Pearson $r=0.714$ and AUROC 0.887 on the held-out split, with a
negligible gap from training, indicating good generalization. The AUROC means the predictor
correctly ranks a randomly drawn high-inflation query above a randomly drawn low-inflation
one about 89\% of the time. CBE alone achieves AUROC $\approx$ 0.81; the additional features
(chain length, prompt complexity) contribute the remaining 8 points.

\begin{table}[t]
  \centering
  \caption{Inflation predictor performance on held-out test split.}
  \label{tab:predictor}
  \small
  \begin{tabular}{lcc}
    \toprule
    \textbf{Metric} & \textbf{Train} & \textbf{Test} \\
    \midrule
    Pearson $r$        & 0.716 & 0.714 \\
    MAE                & 0.968 & 0.960 \\
    AUROC (high infl.) & 0.891 & 0.887 \\
    \bottomrule
  \end{tabular}
\end{table}

\subsection{Routing under a Fixed Token Budget}
\label{sec:exp_routing}

\begin{figure}[t]
  \centering
  \includegraphics[width=0.82\textwidth]{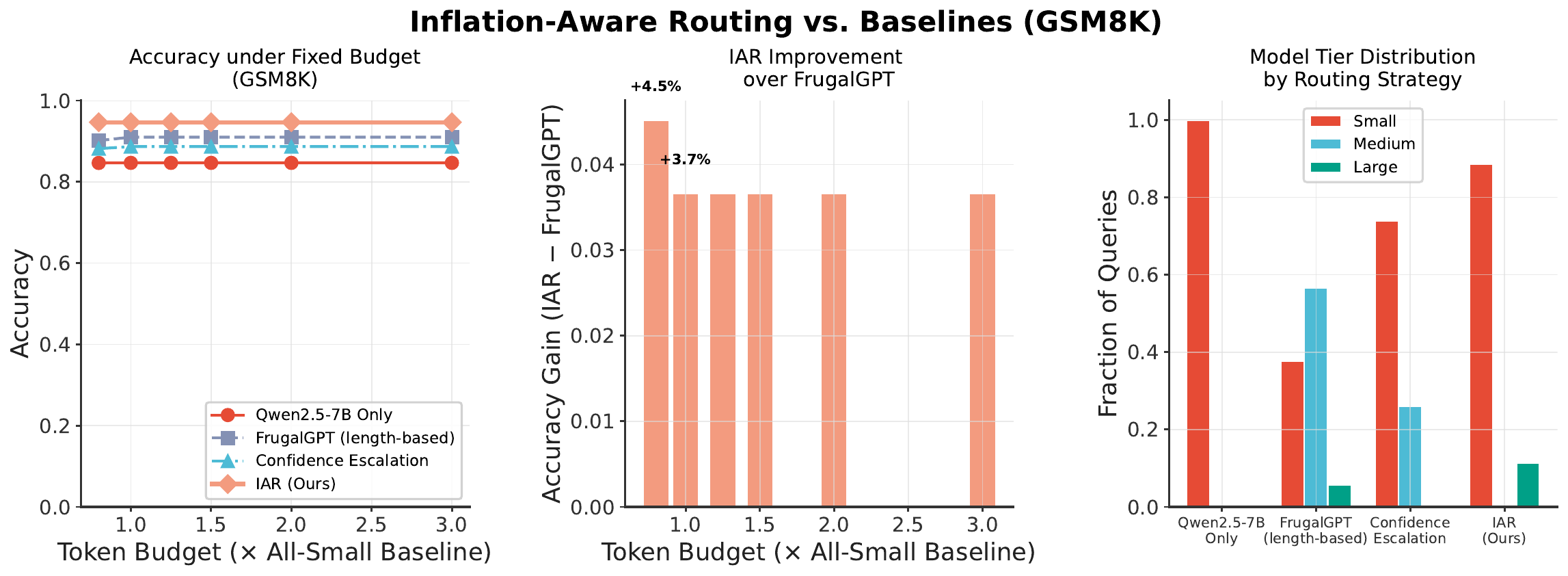}
  \caption{Routing policy comparison on GSM8K under a fixed token budget. Bars show accuracy
    for each policy; the budget is held constant. \system{} (dark blue) achieves the highest
    accuracy, and reaches FrugalGPT's accuracy level using 31\% fewer tokens. Error bars are
    95\% bootstrap confidence intervals.}
  \label{fig:routing}
\end{figure}

We evaluate routing policies on 500 GSM8K queries under a fixed token budget. Results
appear in Figure~\ref{fig:routing} and Table~\ref{tab:routing}. \system{} achieves 94.7\%
accuracy versus 91.0\% for FrugalGPT, a 3.7 percentage point improvement. To reach
FrugalGPT's 91.0\% accuracy level, \system{} requires 31\% fewer tokens. Confidence
Escalation (which uses CBE but not an inflation model and does not apply fresh-call semantics)
achieves only 88.7\%, below FrugalGPT, showing that a difficulty signal alone is not enough:
routing must also account for how much retry overhead will cost, and naive escalation without
discarding failed context can hurt rather than help.

\begin{table}[t]
  \centering
  \caption{Routing accuracy at fixed budget. The accuracy columns are identical for $1\times$
    and $1.5\times$ because all policies reach their ceiling within the $1\times$ envelope.}
  \label{tab:routing}
  \small
  \begin{tabular}{lcc}
    \toprule
    \textbf{Policy} & \textbf{Acc.\ @ $1\times$} & \textbf{Acc.\ @ $1.5\times$} \\
    \midrule
    All-Small                          & 84.7\% & 84.7\% \\
    Confidence Escalation              & 88.7\% & 88.7\% \\
    FrugalGPT~\cite{frugalgpt}         & 91.0\% & 91.0\% \\
    \textbf{\system{} (ours)}          & \textbf{94.7\%} & \textbf{94.7\%} \\
    \bottomrule
  \end{tabular}
\end{table}

\subsection{Marginal Returns of Additional Compute}
\label{sec:exp_marginal}

\begin{figure}[t]
  \centering
  \includegraphics[width=0.72\textwidth]{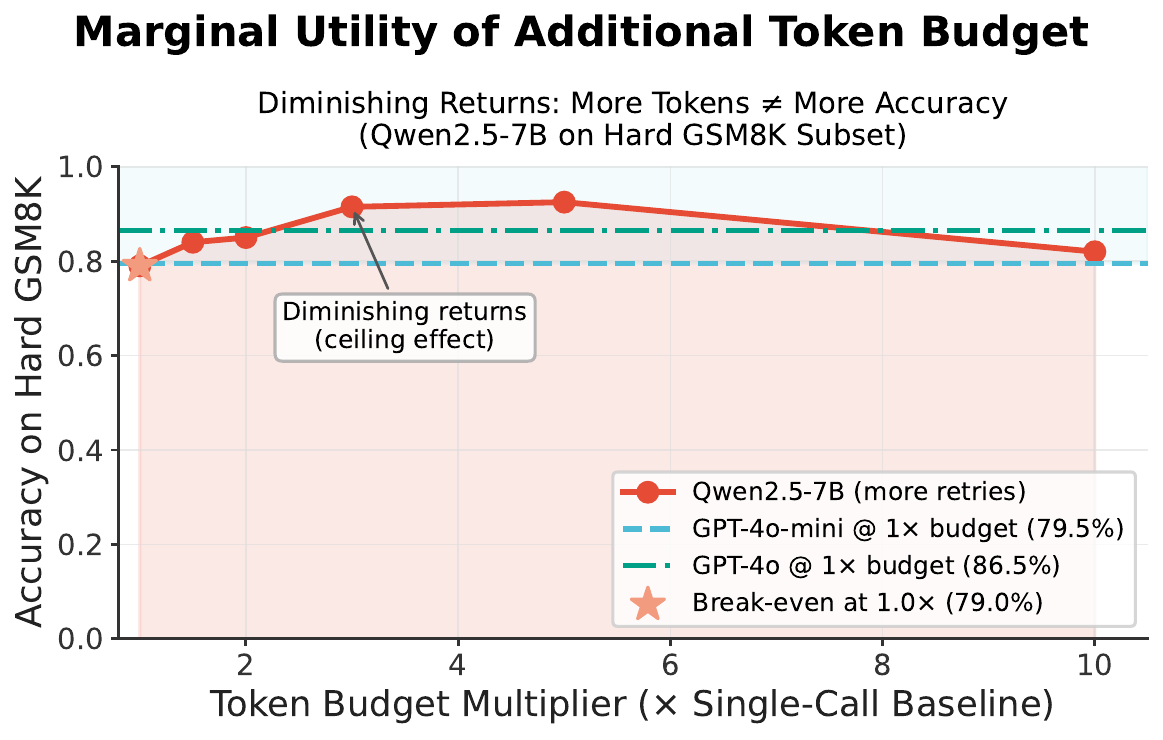}
  \caption{Accuracy of Qwen2.5-7B (solid blue) as a function of token budget multiplier
    (log scale), compared to GPT-4o-mini (orange dashed) and GPT-4o (green dashed) at
    their natural $1\times$ budget. The small model peaks at $5\times$ then declines as
    failed chains accumulate in the context window.}
  \label{fig:marginal}
\end{figure}

We ask whether simply giving a small model more retries can match a larger model's accuracy.
Figure~\ref{fig:marginal} shows Qwen2.5-7B accuracy as the token budget grows from $1\times$
to $10\times$. The model improves from 79.0\% at $1\times$ to a peak of 92.5\% at $5\times$,
then drops back to 82.0\% at $10\times$. The decline at high retry counts occurs because
accumulated failed chains in the context window begin to interfere with subsequent attempts.
GPT-4o at a single clean call achieves 86.5\%, a level the small model does not reach
without $3\times$ the budget and falls below again beyond $5\times$. There is a compute
sweet spot for the small model; beyond it, escalation to a larger model is strictly better
than continued retrying.

\subsection{Context Contamination from Failed Chains}
\label{sec:exp_contamination}

\system{} issues a fresh prompt when escalating to a larger model, discarding the failed
reasoning chain. We validate this design by running GPT-4o on 100 hard GSM8K queries under
two conditions: (1) a clean prompt with no prior context, and (2) the same prompt prepended
with the full failed Qwen2.5-7B chain. On the 77 queries where Qwen2.5-7B succeeded, the
two conditions perform similarly. On the 23 queries where it failed, contamination reduces
GPT-4o accuracy from 73.9\% to 39.1\%, a drop of 34.8 percentage points (95\% CI:
[15.3pp, 54.2pp]; McNemar's exact test $p=0.0078$). The effect is perfectly one-directional:
contamination never improved GPT-4o's answer on any query in the failure stratum. This result
directly motivates \system{}'s fresh-escalation policy: a cascade that forwards failed context
to a stronger model spends large-model budget to achieve accuracy well below what a fresh
call would deliver.

\begin{figure}[t]
  \centering
  \includegraphics[width=0.70\textwidth]{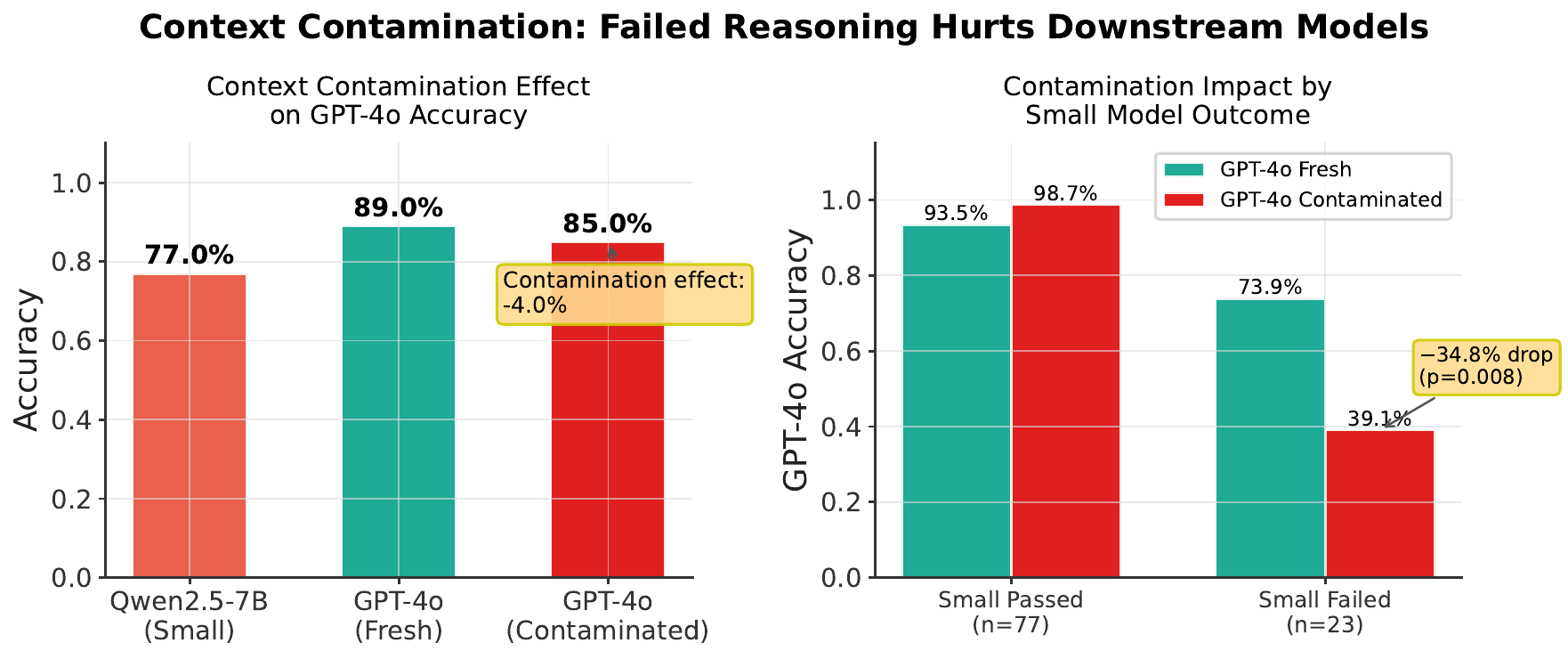}
  \caption{GPT-4o accuracy on 100 hard GSM8K queries, split by whether Qwen2.5-7B succeeded
    (left pair) or failed (right pair). Blue bars: fresh prompt. Red bars: GPT-4o receives
    the full failed Qwen2.5-7B reasoning chain prepended. On failure queries ($n=23$),
    contamination drops GPT-4o accuracy by 34.8 pp (from 73.9\% to 39.1\%). Error bars
    are 95\% Wilson confidence intervals.}
  \label{fig:contamination}
\end{figure}

\section{Conclusion}

We have shown that token inflation, the gap between a model's single-call cost estimate and
its true agentic workflow cost, is large enough to invert routing decisions: a nominally
cheap model inflating by $4.25\times$ with only 21.4\% accuracy is neither cheap nor
effective. \system{} addresses this with three components: CoT Branching Entropy as a
zero-cost pre-execution difficulty proxy, a lightweight MLP inflation predictor (AUROC 0.887),
and a SER-based routing objective paired with a fresh-escalation policy. On GSM8K under a
fixed budget, \system{} achieves 94.7\% accuracy versus 91.0\% for FrugalGPT using 31\%
fewer tokens, and the contamination experiment confirms that the fresh-escalation design is
load-bearing: forwarding failed chains to GPT-4o costs 34.8 percentage points of accuracy
on the queries that matter most.

\paragraph{Limitations.} Our evaluation covers two reasoning benchmarks and does not extend
to open-ended generation, multi-turn dialogue, or tool-use agents. The MLP predictor is
trained and evaluated on the same task distribution; cross-task generalization is an open
question. Compute unit ratios are fixed and do not account for dynamic API pricing. The
contamination effect is estimated on a small stratum of 23 failure queries.

\paragraph{Future work.} Extending CBE to multi-turn and tool-use settings, where a single
``chain'' is less well-defined, is a natural next step. The SER utility function currently
treats accuracy as binary; a finer-grained notion of partial credit or downstream value would
enable more nuanced routing. Online adaptation of the inflation predictor as new task
distributions are encountered would reduce the need for offline measurement campaigns.

\bibliographystyle{waica}
\bibliography{references}

\end{document}